\documentclass[sigconf]{acmart}
\usepackage{pifont}
\usepackage{makecell}
\usepackage{multirow}
\usepackage{graphicx}
\usepackage{xcolor}
\usepackage{empheq}
\usepackage{soul}
\AtBeginDocument{%
	}

\copyrightyear{2026}
\acmYear{2026}
\setcopyright{cc}
\setcctype{by}
\acmConference[MM '26]{Proceedings of the 34th ACM International Conference on Multimedia}{November 10--14, 2026}{Rio de Janeiro, Brazil}
\acmBooktitle{Proceedings of the 34th ACM International Conference on Multimedia (MM '26), November 10--14, 2026, Rio de Janeiro, Brazil}
\acmDOI{10.1145/3767308.3835029}
\acmISBN{979-8-4007-2213-4/2026/11}

\definecolor{subjectRed}{HTML}{FF0000}
\definecolor{motionOrange}{HTML}{C55A11}
\definecolor{objectBlue}{HTML}{0070C0}
\definecolor{eqGray}{HTML}{D8D8D8}

\begin{document}
	
	\title{Knowledge-guided Disentanglement with Atomic Actions for Action Recognition}
	
	\makeatletter
	\let\old@fnsymbol\@fnsymbol
	\renewcommand{\@fnsymbol} [1]{%
		\ifnum#1=1
		\ensuremath{\dagger}%
		\else
		\old@fnsymbol{#1}%
		\fi}
	\makeatother 
	
	\author{Tianci Wu}
	\affiliation{
		\institution{Xidian University}
		\city{Xi'an}
		\country{China}}
	\email{tianciwu@stu.xidian.edu.cn}

		\author{Siqi Cao}
	\affiliation{
		\institution{Xidian University}
		\city{Xi'an}
		\country{China}}
	\email{25031212111@stu.xidian.edu.cn}

	\author{Guangming Zhu}
	\authornote{ Corresponding authors}
	\affiliation{
		\institution{Xidian University}
		\city{Xi'an}
		\country{China}}
	\email{gmzhu@xidian.edu.cn}

	\author{Jiang Lu}
\affiliation{
	\institution{Xidian University}
	\city{Xi'an}
	\country{China}}
\email{24031111038@stu.xidian.edu.cn}

	\author{Siyuan Wang}
\affiliation{
	\institution{Xidian University}
	\city{Xi'an}
	\country{China}}
\email{siyuanwang@stu.xidian.edu.cn}

	\author{Longfei Zhang}
\affiliation{
	\institution{National University of Defense Technology}
	\city{Changsha}
	\country{China}}
\email{zhanglongfei@nudt.edu.cn}

	\author{Jincai Huang}
\affiliation{
	\institution{Hunan Institute of Advanced Technology}
	\city{Changsha}
	\country{China}}
\email{huangjincai@nudt.edu.cn}

	\author{Jun Sheng}
\affiliation{
	\institution{Shanghai Road Transport Development Center}
	\city{Shanghai}
	\country{China}}
\email{shengjun0325@126.com}

	\author{Liang Zhang}
\affiliation{
	\institution{Xidian University}
	\city{Xi'an}
	\country{China}}
\email{liangzhang@xidian.edu.cn}


	
	\begin{abstract}

		Action recognition in complex scenes often involves multiple concurrent fine-grained actions, making it challenging to model internal action structures. Most existing methods rely on holistic representations, which are insufficient for capturing subtle interactions and fine-grained semantics. While recent prompt-based approaches introduce disentanglement, they lack explicit semantic guidance, and methods based solely on visual or structured cues remain coarse-grained. In this paper, we propose Knowledge-guided Disentanglement with Atomic Actions (KDA), which leverages fine-grained semantic knowledge to enhance action representations and enable more precise disentanglement. Specifically, we use Large Language Models (LLMs) to decompose action labels into atomic actions, providing explicit spatial-temporal semantics. A Knowledge Injection Module (KIM) first integrates atomic action knowledge into video features. Based on this enhanced representation, a Knowledge Disentanglement Module (KDM) further disentangles atomic action knowledge to produce more precise semantic guidance for action disentanglement. A Knowledge Disentanglement Loss (KD Loss) is introduced to encourage clearer disentanglement of knowledge components within KDM. Extensive experiments demonstrate that KDA improves feature discriminability and achieves state-of-the-art performance on multi-label action recognition benchmarks. Moreover, KIM and KDM can be readily integrated into other methods, demonstrating strong generality. The code is available \href{https://github.com/iamsnaping/ProDA}{\textcolor{blue}{here}}.
	\end{abstract}



	\maketitle
	\vspace{-10pt}
	\section{Introduction}
	Action recognition is a fundamental task in video understanding~\cite{hassner2013critical}. Particularly in complex scenarios (e.g., multiple interacting targets and concurrent actions), it remains highly challenging. Most existing methods encode video clips as holistic representations using architectures such as 2D Convolutional Neural Networks (CNNs)~\cite{feichtenhofer2017spatiotemporal,feichtenhofer2016convolutional,simonyan2014two}, 3D CNNs~\cite{tran2015learning,carreira2017quo,xie2018rethinking}, or Vision Transformers (ViTs)~\cite{wang2024omnivid,li2022mvitv2,liu2022video,arnab2021vivit}. However, in complex scenes, multiple objects and their interactions are highly entangled, making holistic representations insufficient for modeling object interactions. ProDA~\cite{proda} introduces a disentanglement perspective by leveraging learnable prompts to separate action representations from complex video scenes. However, purely learnable prompts contain limited explicit semantic information, struggling to sufficiently enlarge the feature margin between similar actions. Moreover, relying solely on visual cues, these approaches often struggle to produce highly discriminative representations, hindering effective action understanding. These limitations highlight the need for richer and more informative semantic cues to enhance feature discriminability and better capture complex interactions in video understanding. In this paper, we build upon the disentanglement perspective and aim to enhance it by incorporating richer semantic guidance.
	
	\begin{figure}[!t]
		\centering
		\includegraphics[width=1.00\linewidth]{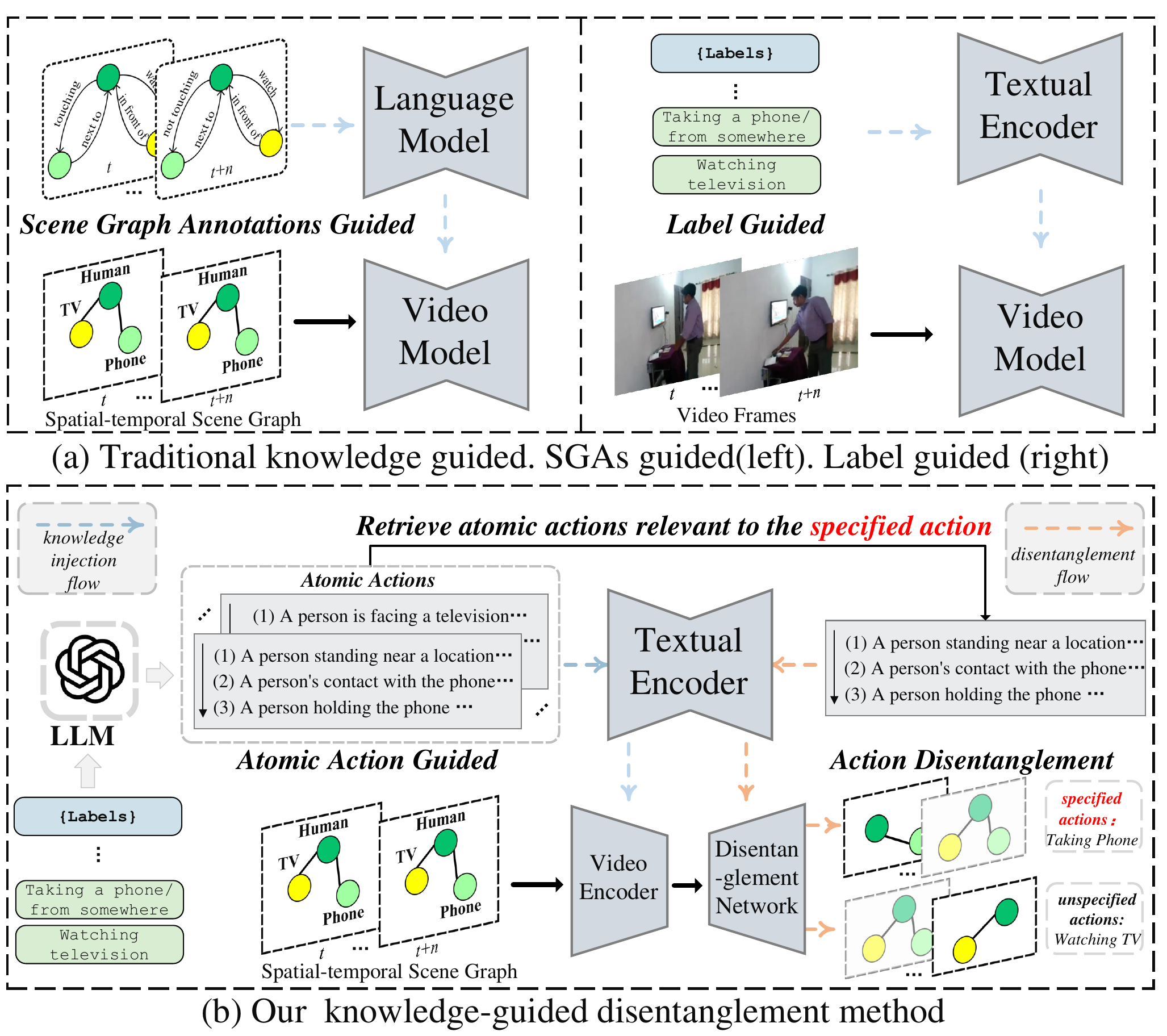}
		\caption{(a) Traditional knowledge-guided paradigms based on scene graph annotations (left) or coarse labels (right). (b) Our framework: LLM-generated atomic actions provide fine-grained knowledge to enhance SSG features and guide disentanglement into specified and unspecified actions.}
		\label{fig:first}
		\vspace{-5pt}
	\end{figure}

	Many methods improve action discriminability by introducing structured or semantic cues, yet remain limited in semantic granularity. Graph-based methods represent videos as structured graphs and process them with ViTs or GNNs~\cite{zhang2024object,materzynska2020something,zhou2023can,wang2018videos,zhang2020temporal,zhang2019structured}. Some works construct Spatial-temporal Scene Graphs (SSGs) to model dynamic interactions, often relying on scene graph annotations (SGAs)~\cite{zhuo2019explainable,actiongenome,lair,ou2022object,jin2022complex}. However, they still lack explicit semantic guidance, and SGAs are limited in adapting to dynamic actions. Language-enhanced methods introduce external knowledge by converting structured representations into textual forms and leveraging language models (LMs) for reasoning~\cite{lair}, yet such knowledge is typically constrained to entity-level descriptions and coarse relations~\cite{alam2022language}. In parallel, vision-language approaches based on models such as CLIP~\cite{clip} inject label semantics into video representations~\cite{victr,wang2109actionclip,luo2021clip4clip}, but remain restricted by coarse-grained label annotations. Overall, these approaches either lack explicit semantic guidance or fail to effectively integrate rich semantics with structured visual representations, limiting their ability to model complex interactions. To address these limitations, recent works~\cite{actalign,qian2025beyond} utilize Large Language Models (LLMs) to generate atomic actions from labels, which encode both spatial and temporal information. Inspired by this, \textit{\textbf{we argue that injecting atomic action knowledge into SSGs can improve action discriminability and facilitate more effective disentanglement.}} However, this process faces two key challenges: (1) the lack of node-level supervision in SSGs makes it unclear how to associate atomic actions with graph nodes, especially under multi-action scenarios; (2) atomic actions and learnable prompts exhibit different limitations: atomic actions may introduce overly strong semantics that overwhelm visual representations, while learnable prompts often provide limited semantic expressiveness, making effective disentanglement challenging.

	To this end, we propose Knowledge-guided Disentanglement with Atomic Actions for Action Recognition (KDA). As illustrated in Fig.~\ref{fig:first}(c), we first inject atomic actions containing rich spatial-temporal knowledge into SSGs to enhance feature discriminability. Based on the enhanced representations, we further leverage atomic actions to guide action disentanglement from video representations.~\textbf{Our framework performs hierarchical knowledge disentanglement at feature, structure, and semantic levels. }Specifically, at the feature level, we introduce a Knowledge Injection Module (KIM) that adaptively associates atomic action knowledge with SSG nodes. At the structure level, a Knowledge Disentanglement Module (KDM) leverages learnable embeddings and atomic knowledge to generate precise semantic guidance for action disentanglement. At the semantic level, we further introduce a Knowledge Disentanglement Loss (KD Loss) to encourage discriminative and independent semantic cues within KDM.
	
	In summary, the major contributions of this work are as follows:
	\begin{itemize}
		\item We propose a hierarchical knowledge disentanglement framework across feature, structural, and semantic levels for modeling complex multi-action scenarios.
		\item Our method achieves state-of-the-art performance on multi-label action recognition benchmarks, highlighting its ability to capture complex and diverse action semantics.
		\item We validate the generality of our approach by integrating KIM and KDM into different baselines, consistently improving performance.

	\end{itemize}
	
	\section{Related Works}
	Existing action recognition methods can be broadly categorized into CNN-based, ViT-based, object-centric, and VLM-based methods, focusing on increasingly rich spatial-temporal and semantic modeling of video content.
	\subsection{Action Recognition}
	Existing action recognition methods can be broadly categorized into CNN-based, ViT-based, and object-centric methods. CNN-based methods extend 2D CNNs with temporal modeling, including two-stream networks and 3D CNNs~\cite{feichtenhofer2017spatiotemporal,feichtenhofer2016convolutional,simonyan2014two,wang2013action,wang2017spatiotemporal,tran2015learning,carreira2017quo,xie2018rethinking}. ViT-based methods model videos as sequences of tokens using Transformers~\cite{girdhar2019video,wang2024omnivid,li2022mvitv2,liu2022video}, but often struggle with noisy holistic representations. Object-centric methods leverage regions of interest or scene graphs to model interactions~\cite{zhang2024object,materzynska2020something,zhou2023can,wang2018videos,zhuo2019explainable,actiongenome}, yet still process all actions jointly and have limited capability in multi-action scenarios. ProDA~\cite{proda} introduces prompt-based disentanglement using learnable prompts to extract specified actions from video representations. However, learnable prompts provide limited semantic guidance. In contrast, we first leverage LLM-generated atomic actions to enhance video features, then use the atomic actions to provide richer semantics to guide action disentanglement.
	\subsection{VLM-based Video Recognition}
	Adapting VLMs to video tasks has recently attracted increasing attention. Several works extend image-based VLMs to video understanding by incorporating temporal modeling, multi-modal fusion, or adapting pretrained encoders through prompt learning or fine-tuning~\cite{cheng2023vindlu,qian2022multimodal,ju2022prompting,rasheed2023fine,xue2022clip,yu2022coca}. Among them, a line of work focuses on extending CLIP~\cite{clip} for video understanding. ActionCLIP~\cite{wang2109actionclip} and CLIP4Clip~\cite{luo2021clip4clip} introduce temporal modules to model frame relationships, while EVL~\cite{lin2022frozen}, and X-CLIP~\cite{ni2022expanding} explore different strategies for adapting CLIP to video tasks. VicTR~\cite{victr} further improves video-text alignment through interactions between visual and textual representations. However, these methods mainly rely on action labels as textual guidance, which cannot explicitly model interactions between objects In contrast, our method operates on SSG representations and injects atomic actions generated by LLMs as knowledge guidance, enabling better action understanding.

	\subsection{Disentangled Representation Learning}
	Existing disentangled representation learning (DRL) methods aim to separate underlying factors of variation from different perspectives.  Some works focus on attribute-level disentanglement, while others decompose representations into shared and private components, often using contrastive learning objectives~\cite{PSGAN,DARLING,hsieh2018learning,denton2017unsupervised,zhu2020s3vae,gan2025dfdnet,mo2023disentangled}. In structured domains, several approaches further extend disentanglement to graphs by decomposing them into latent components or subgraphs~\cite{graph_dis,li2022disentangled,li2021disentangled}. However, these methods are primarily designed for static or single-factor decomposition and struggle to handle dynamic scenarios with multiple entangled actions. 
	ProDA~\cite{proda} extends disentanglement to action recognition by employing learnable prompts to guide the targeted disentanglement of arbitrarily specified actions from complex scenes. However, such prompts provide limited semantic information and are insufficient for disentangling complex action dynamics. In contrast, our method leverages atomic actions containing fine-grained spatial-temporal information to achieve more precise action disentanglement.
	
	\begin{figure*}[!t]
		\centering
		\includegraphics[width=1.00\textwidth]{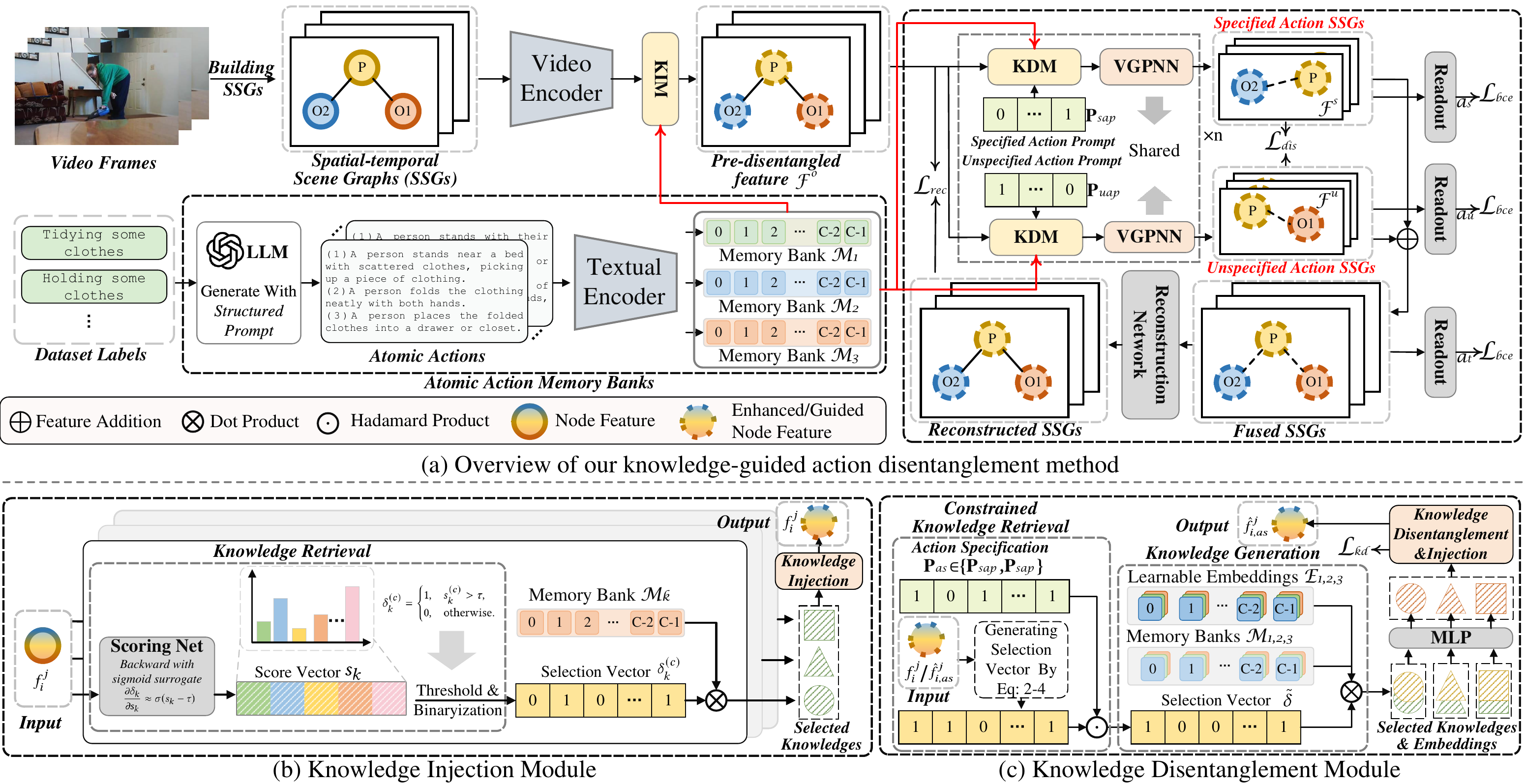}
		\caption{(a) An overview of our method. We first construct SSGs and extract node features using a video encoder. Meanwhile, action labels are decomposed into atomic actions via LLMs and stored in memory banks. These atomic actions are further used to enhance video features and guide action disentanglement. (b) The KIM adaptively retrieves and aggregates atomic knowledge to enhance node representations. The AS defines the target actions for disentanglement. (c) Conditioned on the AS, the KDM disentangles atomic actions relevant to the specified actions to generate more precise semantic cues.}
		\label{fig:overall}
		\vspace{-5pt}
	\end{figure*}
	
	\section{Methodology}
	\subsection{Framework}

	As illustrated in Fig.~\ref{fig:first}(c), our method leverages atomic actions to enhance video features and guide action disentanglement in complex multi-action scenarios. Specifically, we incorporate atomic action priors to enrich feature representations and employ fine-grained atomic cues to guide precise action disentanglement. The overall framework is shown in Fig.~\ref{fig:overall}. Given an input video, we first represent it as an SSG. Meanwhile, action labels are decomposed into atomic actions via an LLM with structured prompts, forming \textbf{Atomic Action Memory Banks (AMBs)}.  The SSGs are encoded by~\cite{han2022dual} to capture spatial-temporal interactions, followed by a Knowledge Injection Module (KIM) that incorporates atomic knowledge to enhance node features~(\S \ref{sec:kim}). A Knowledge Disentanglement Module (KDM) then generates semantic guidance from atomic knowledge to facilitate action disentanglement~(\S \ref{sec:kdm}). Finally, the guided features are fed into a Video Graph Parsing Neural Network (VGPNN)~\cite{proda} for action disentangling, with a Knowledge Disentanglement Loss ($\mathcal{L}_{kd}$) to enforce discriminative and independent semantic cues (\S \ref{sec:kdl}).

	\textbf{Formulation}. In an SSG, each video frame is represented as a graph, where nodes correspond to objects and edges describe semantic relationships between them (e.g., spatial or interaction relations). Formally, we denote the SSG of a video as $\mathcal{G} = (\mathcal{V}, \mathcal{R})$, where $\mathcal{V}$ is the set of object nodes and $\mathcal{R}$ is the set of relations. For each frame $i$, we denote its node set as $\mathcal{V}_i$, and $v_i^j$ represents the $j$-th object in frame $i$. The node features are denoted as $\mathcal{F}_{\text{node}} \in \mathbb{R}^{N \times M \times D}$, where $N$, $M$, and $D$ indicate the number of frames, objects per frame, and feature dimension, respectively.

	\textbf{Atomic Action Memory Bank}. To leverage rich prior knowledge, we decompose each action label into a sequence of fine-grained atomic action descriptions using LLMs, following the prompt design in~\cite{qian2025beyond}. These descriptions capture structured spatial-temporal semantics of actions. As shown in Fig.~\ref{fig:overall} (a), the atomic actions are encoded by a textual encoder to obtain embeddings $\{t_k\}_{k=1}^{K}$, where $t_k \in \mathbb{R}^{D}$ denotes the embedding of the $k$-th atomic action. Given $C$ action classes, we organize these embeddings into a set of atomic knowledge memory banks $\{\mathcal{M}_k\}_{k=1}^{K}$, where each memory bank is defined as:
	\begin{equation}\small
		\mathcal{M}_k = \{t_k^{(c)}\}_{c=1}^{C} \in \mathbb{R}^{C \times D}.
	\end{equation}
	This formulation provides structured atomic knowledge for subsequent retrieval and injection.
	\vspace{-2pt}
	\subsection{Knowledge Injection Module (KIM)}
	\label{sec:kim}
	The Knowledge Injection Module (KIM) injects atomic action knowledge to enhance action representations. In multi-label scenarios, each node in the SSGs may participate in a variable number of overlapping actions. To accommodate this, we adopt an adaptive knowledge injection mechanism that dynamically selects a variable number of atomic actions per node. Specifically, we first perform \textbf{knowledge retrieval} to score atomic actions from AMBs. To enable gradient propagation through discrete selection, we employ a sigmoid-based surrogate within the Straight-Through Estimator (STE)~\cite{ste}. Finally, the selected atomic knowledge is aggregated and injected into the node representation via \textbf{knowledge injection}.
	
	\textbf{Knowledge Retrieval.} Given a node feature $f_i^j$, we compute a relevance score for each atomic knowledge memory bank. For the $k$-th memory bank $\mathcal{M}_k$, we introduce a scoring network $\phi_k(\cdot)$ that produces a score vector
	\begin{equation}\small
		\label{eq:score}
		s_k = \phi_k(f_i^j), \quad s_k \in \mathbb{R}^{C}.
	\end{equation}
	
	The scores are then converted into binary selection indices using a threshold $\tau$. For the $k$-th memory bank, we obtain a binary selection vector $\delta_k \in \{0,1\}^{C}$, where each element indicates whether the corresponding atomic knowledge entry is selected:
	\begin{equation}\small
		\label{eq:threshold}
		\delta_k^{(c)} =
		\begin{cases}
			1, & s_k^{(c)} > \tau, \\
			0, & \text{otherwise}.
		\end{cases}
	\end{equation}
	
	Since the hard thresholding operation is non-differentiable, we employ the straight-through estimator (STE)~\cite{ste} to facilitate gradient propagation. Specifically, during backpropagation, we approximate the gradient using a sigmoid surrogate:
	\begin{equation}\small
		\label{eq:ste}
		\frac{\partial \delta_k}{\partial s_k} \approx \sigma(s_k - \tau),
	\end{equation}
	where $\sigma(\cdot)$ denotes the sigmoid function. This design stabilizes training by avoiding gradient explosion, while preserving non-zero gradients for suppressed entries, thus maintaining the model’s ability to explore informative components.

	\textbf{Knowledge Injection.} Given the selection indices, atomic knowledge is retrieved from each memory bank via matrix multiplication, formulated as $z_k = \delta_k^\top \mathcal{M}_k$. To prevent semantic imbalance caused by varying knowledge densities across different memory banks, we concatenate the retrieved knowledge vectors and employ a Multi-Layer Perceptron (MLP) for adaptive aggregation. Finally, the aggregated knowledge is injected into the node representation through residual fusion:
	\begin{equation}\small
		\label{eq:injection}
		f_i^j \leftarrow f_i^j + \mathtt{MLP}([z_1 ; z_2 ; \dots ; z_K]),
	\end{equation}
	where $[;]$ denotes the concatenation operation. This residual injection yields an \textbf{enhanced node feature}, serving as a robust foundation for the subsequent structural disentanglement.

	\subsection{Knowledge Disentanglement Module (KDM)}
	\label{sec:kdm}

	The Knowledge Disentanglement Module (KDM) performs prompt-conditioned action disentanglement using Action Specification (AS), AMBs, and learnable embeddings. Unlike KIM, it emphasizes semantic guidance, where AS constrain knowledge retrieval toward specified actions and embeddings balance semantic dominance. Specifically, KDM performs \textbf{constrained knowledge retrieval}, followed by \textbf{knowledge generation} and \textbf{knowledge disentanglement \& injection}, whose resulting components are aggregated to guide action disentanglement.
	
	\textbf{Action Specification.} To guide action disentanglement, we define an Action Specification (AS) that encodes specified actions. Following ProDA~\cite{proda}, each action is represented as a one-hot vector, and AS is constructed as a multi-hot vector. It consists of two complementary components: the Specified Action Prompt (SAP), $\mathbf{p}_{sap} \in \{0,1\}^C$, indicating specified actions, and the Unspecified Action Prompt (UAP), $\mathbf{p}_{uap} = \mathbf{1} - \mathbf{p}_{sap}$, capturing the remaining actions. This design explicitly models both target and residual semantics while mitigating information leakage. To improve robustness, we adopt random sampling~\cite{proda} to generate UAPs and SAPs. Since both prompts are processed identically in KDM, we use a unified notation $\mathbf{p}_{as} \in \{\mathbf{p}_{sap}, \mathbf{p}_{uap}\}$.
	
	\textbf{Constrained Knowledge Retrieval.} Given an enhanced node feature $f_i^j$, we compute a base selection vector $\delta^{(c)} \in \{0,1\}^{C}$ following the identical scoring and STE thresholding process described in KIM (Eq.~\ref{eq:score}-\ref{eq:ste}). However, since KDM performs conditional disentanglement, we explicitly constrain the selection index using the prompt mask to ensure only specified knowledge is preserved:
	\begin{equation}\small
		\label{eq:mask}
		\tilde{\delta} = \delta \odot \mathbf{p}_{as},
	\end{equation}
	where $\odot$ denotes the Hadamard product. The constrained knowledge is then retrieved from the memory banks $\tilde{z_k} = \tilde{\delta}^\top \mathcal{M}_k$, ensuring that the extracted knowledge is strictly aligned with the given action specification $\mathbf{p}_{as}$.

	\textbf{Knowledge Generation.} When relying solely on retrieved knowledge for guidance, the strong semantic priors may dominate the representation and suppress the original video information. To address this issue, inspired by~\cite{darcet2023vision}, we introduce an additional learnable embedding for each memory bank. Specifically, we equip each memory bank with a parallel learnable matrix $\mathcal{E}_k \in \mathbb{R}^{C \times D}$, which shares the same dimensionality as $\mathcal{M}_k$ and provides extra adaptive capacity beyond the knowledge. We extract the learnable embedding $e_k \in \mathbb{R}^D$ using the same constrained selection vector, i.e., $e_k = \tilde{\delta}^\top \mathcal{E}_k$. The retrieved knowledge $z_k$ and the corresponding learnable embedding $e_k$ are then concatenated and fused by MLP.
	\begin{equation}\small
		\label{eq:retrieval_kdm}
		\hat{z}_k = \mathtt{MLP}([z_k ; e_k]).
	\end{equation}
	We then aggregate the representations from all $K$ memory banks as $\hat{z} = [\hat{z}_1 ; \dots ; \hat{z}_K]$.
	
	\textbf{Knowledge Disentanglement \& Injection.}
	Since the retrieved knowledge entangles multiple atomic action semantics, we employ $M$ disentanglement heads to decompose it into more fine-grained semantic components.
	\begin{equation}\small
		h_m = \psi_m(\hat{z}),
	\end{equation}
	where $\psi_m(\cdot)$ denotes the $m$-th disentanglement head.

	To aggregate the disentangled components, we predict dynamic weights $\alpha \in \mathbb{R}^{M}$ conditioned on the node feature, accounting for their varying relevance:
	\begin{equation}\small
		\label{eq:guidance}
		g_{as} = \sum_{m=1}^{M} \alpha_m h_m, \quad \text{where} \;\; \alpha = \mathtt{MLP}([f_i^j ; h_1 ; \dots ; h_M]).
	\end{equation}
	
	This design allows each node to selectively aggregate the most relevant disentangled knowledge components. Detailed analyses of the selection strategy and ablations on the number of disentanglement heads are provided in the supplementary material.
	
	By separately feeding $\mathbf{p}_{sap}$ and $\mathbf{p}_{uap}$ into this process, we obtain $g_{sap}$ and $g_{uap}$. Finally, these signals are independently fused with the original node features, yielding \textbf{guided node features}:
	\begin{equation}\small
		\label{eq:fusion}
		\hat{f}_{i, as}^j = \mathtt{MLP}([f_i^j ; g_{as}]), \quad as \in \{SAP, UAP\}.
	\end{equation}
	This results in two complementary, explicitly guided feature spaces ($\hat{f}_{i, SAP}^j$ and $\hat{f}_{i, UAP}^j$), which are subsequently fed into the VGPNN for relational reasoning.

	\subsection{Optimization}

	In this section, we present a hierarchical loss formulation operating at the feature, structural, and semantic levels. At the feature level, a classification loss supervises representation learning and enforces alignment with the given AS. At the structural level, an Action Disentanglement (AD) Loss promotes disentanglement between specified and unspecified actions (i.e., remaining actions), enabling clearer disentanglement of action-relevant interactions. At the semantic level, a Knowledge Disentanglement (KD) Loss ($\mathcal{L}_{kd}$) regularizes disentanglement heads in KDM, encouraging mutual independence while preserving fine-grained semantic knowledge.
	These objectives operate in a coarse-to-fine manner, progressively improving representation quality and facilitating structured disentanglement in multi-action scenarios.
	
	\textbf{Action Classification Loss.} 
	To ensure consistency with the Action Specification and preserve information completeness, we apply a standard Binary Cross-Entropy (BCE) loss to the predictions derived from the UAP-guided ($a_u$), SAP-guided ($a_s$), and fused representations ($a_t$). 
	These predictions are obtained by applying a readout function (implemented by MLP) over the corresponding graph features.
	Given their corresponding ground-truth labels $\{y_u, y_s, y_t\}$, the classification objective is formulated as:
	\begin{equation}\small
		\mathcal{L}_{cls} = \mathcal{L}_{bce}(a_u, y_u) + \mathcal{L}_{bce}(a_s, y_s) + \mathcal{L}_{bce}(a_t, y_t).
	\end{equation}

	\textbf{Action Disentanglement Loss (AD Loss).} To facilitate action disentanglement, we adopt the Action Disentanglement Loss~\cite{proda}, defined as $\mathcal{L}_{ad} = \mathcal{L}_{dis} + \mathcal{L}_{rec}$. $\mathcal{L}_{dis}$ minimizes the Pearson correlation ($\mathtt{Corr}$) between the UAP-guided ($\mathcal{F}^u$) and SAP-guided ($\mathcal{F}^s$) representations to encourage orthogonality, while $\mathcal{L}_{rec}$ reconstructs the original pre-disentangled feature $\mathcal{F}^o$ to preserve fidelity. We omit the batch index and present the objectives as follows:
	\begin{equation}\small
		\label{eq:ad_loss}
		\begin{aligned}
			\mathcal{L}_{dis} &= \mathtt{ReLU} \left( \mathtt{Corr}(\phi_{dis}(\mathcal{F}^{u}),\psi_{dis}(\mathcal{F}^{s})) - m_{1} \right), \\
			\mathcal{L}_{rec} &= \mathtt{ReLU} \left( \parallel \mathcal{F}^{o} - \mathtt{Net}_{rec}(\mathcal{F}^u, \mathcal{F}^s) \parallel^2_2 - m_{2} \right),
		\end{aligned}
	\end{equation}
	where $m_1$ and $m_2$ are margins preventing over-penalization and representation collapse, $\phi_{dis}$ and $\psi_{dis}$ are measurable functions~\cite{gretton2005kernel}, and $\mathtt{Net}_{rec}$ represents the reconstruction network.
	
	\label{sec:kdl}
	\textbf{Knowledge Disentanglement Loss (KD Loss).} While the AD Loss enforces macro-level disentanglement between specified and unspecified actions, we further introduce a Knowledge Disentanglement Loss (KD Loss) to regularize KDM at the micro\textbf{-}level. Unlike conventional contrastive learning that focuses on representation disentanglement, the proposed KD Loss explicitly enforces both independence and specificity among disentangled knowledge components. To preserve the semantic structure of multiple disentanglement heads, we model the representations into two complementary parts: \textbf{\textit{shared information}} captured across heads, and \textbf{\textit{specific information}} unique to each component. This decomposition prevents the heads from collapsing into redundant shared semantics or drifting away from meaningful category-specific representations.
	
	Based on this formulation, we formulate the KD Loss as a margin-based contrastive learning objective. Let $H = \{\hat{h}_1, \hat{h}_2, \dots, \hat{h}_M\}$ denote the outputs of the $M$ disentanglement heads, and $\mathcal{U} = \{\mu_1, \mu_2, \dots, \mu_L\}$ denote a set of $L$ learnable semantic anchors. We employ these learnable semantic anchors to explicitly anchor the \textbf{\textit{shared information}}, maintaining the foundational knowledge structures. Meanwhile, to preserve the \textbf{\textit{specific information}}, we introduce a margin $\gamma$ that bounds the similarity between the $m$-th head and the $l$-th anchor:
	\begin{equation}\small
		s(\hat{h}_m, \mu_l) = \min(\hat{h}_m^\top \mu_l, \gamma).
	\end{equation}
	By clipping the similarity, the margin $\gamma$ deliberately prevents the heads from completely collapsing into the generic anchors. This bounded variation effectively preserves the inherent, fine-grained details specific to each knowledge head.
	
	To enforce holistic disentanglement, we treat the semantic anchors $\mathcal{U}$ as positive targets to consolidate the shared information, while treating the outputs of other heads $h_n \ (n \neq m)$ as negative samples to enforce inter-head mutual independence. The total KD Loss is formulated by averaging this margin-based contrastive objective across all $M$ heads:
	\begin{equation}\small
		\label{eq:kd_loss}
		\mathcal{L}_{kd} = - \frac{1}{M} \sum_{m=1}^{M} \log \frac{ \sum_{l=1}^L \exp(s(\hat{h}_m, \mu_l) / t) }{ \sum_{l=1}^L \exp(s(\hat{h}_m, \mu_l) / t) + \sum_{n \neq m} \exp({\hat{h}_m}^\top h_n / t) },
	\end{equation}
	where $t$ is the temperature hyperparameter. Minimizing this loss explicitly pushes each head $h_m$ away from the others $h_n$, thereby promoting mutual independence among disentangled knowledge components. This mechanism enables KDM to maintain cohesive shared representations while preserving discriminative, component-specific details. Furthermore, it explicitly enforces structured disentanglement rather than general representation disentanglement.

	\textbf{Overall Objective.} Finally, the overall objective function to train our knowledge-guided disentanglement framework is formulated by combining the classification loss, the AD Loss (for macro-level feature disentanglement), and our proposed KD Loss (for micro-level prior disentanglement):
	\begin{equation}\small
		\mathcal{L}_{total} = \mathcal{L}_{cls} + \lambda_1 \mathcal{L}_{ad} + \lambda_2 \mathcal{L}_{kd},
	\end{equation}
	where $\lambda_1$ and $\lambda_2$ are balancing hyperparameters. This joint optimization strategy ensures accurate multi-label action recognition while achieving precise, fine-grained feature disentangling.

	\section{Experiment}
	\subsection{Datasets and Evaluation Metrics}
	The Charades dataset~\cite{charades} contains 9,848 videos with an average duration of 30 seconds, covering 157 action categories. Each video includes an average of 6.8 actions, and multiple actions may occur simultaneously, making the recognition task particularly challenging. The Action Genome dataset~\cite{actiongenome}, built upon Charades, provides fine-grained annotations by decomposing actions and focusing on video segments where the actions occur. It contains 234K keyframes with annotations for 476K object bounding boxes and 1.72M object relationships. The SportsHHI dataset~\cite{sportshhi} focuses on human-human interactions in sports scenarios. It includes basketball and volleyball videos selected from MultiSports~\cite{li2021multisports}. The dataset defines 34 interaction categories and provides 118,075 human bounding boxes with 50,649 interaction instances. 
	
	Following~\cite{lair,sportshhi}, we evaluate our method on Charades for multi-label action recognition and on SportsHHI for human-human interaction classification (HHICls). The performance is measured using mean Average Precision (mAP).
	\vspace{-3pt}
	\subsection{Experiment Setup}
	\textbf{Charades.} All experiments follow the same training protocol, using $N$ frames per video with random sampling during training and uniform sampling during inference. We evaluate under standard and oracle protocols, where SGAs are obtained from model predictions and ground truth, respectively. Under the standard protocol, we use LaIAR~\cite{lair} visual features, while under the oracle protocol visual features are extracted using CLIP (B/16)~\cite{clip}. In both settings, the textual encoder is CLIP (B/16).
	
	\noindent\textbf{SportsHHI.} On SportsHHI, we adopt the official open-source model and training code without modifying the original training settings~\cite{sportshhi}. We use a CLIP (B/16) textual encoder to encode atomic actions. For evaluation, we conduct two experiments: (1) extending the official baseline by incorporating our proposed modules (e.g., KIM, KDM and KD Loss); and (2) constructing a parameter-matched baseline by enlarging the model size to match ours, without introducing any proposed components.

	\begin{table}
		\centering
		\caption{Multi-label action recognition performance comparison on the Charades dataset in term of mAP. SSG: ground-truth SSG. Bbox: bounding box from ground truth SSG. * denotes that the backbone model is frozen during training.}
		\label{table:charades}
		\centering
		\resizebox{\linewidth}{!}{
			\begin{tabular}{llll}
				\toprule
				Methods&Backbone&Modality&mAP\\
				\midrule
				I3D~\cite{carreira2017quo}&R101-I3D&RGB&15.6\\
				VideoMLN~\cite{jin2022complex}&R101&RGB&38.4\\
				STRG~\cite{wang2018videos}&R101-I3D-NL&RGB&39.7\\
				SGFB~\cite{actiongenome}&R101-I3D-NL&RGB&44.3\\
				OR2G~\cite{ou2022object}&R101-I3D-NL&RGB&44.9\\
				LaIAR~\cite{lair}&R101-I3D-NL&RGB&45.1\\
				ProDA~\cite{proda}&R101&RGB&50.2\\
				KDA (Ours)&R101&RGB&49.9\\
				\midrule
				SGRB Oracle~\cite{actiongenome}&R101-I3D-NL&RGB+SSG&60.3\\
				VideoMLN Oracle~\cite{jin2022complex}&R101-I3D-NL&RGB+SSG&62.8\\
				OR2G Oracle~\cite{ou2022object}&R101&RGB+SSG&63.3\\
				LaIAR Oracle~\cite{lair}&R101&RGB+Bbox&63.6\\
				OR2G Oracle~\cite{ou2022object}&R101-I3D-NL&RGB+SSG&67.5\\
				LaIAR Oracle~\cite{lair}&R101-I3D-NL&RGB+Bbox&67.4\\
				\midrule
				VicTR~\cite{victr} &CLIP (B/16) & RGB & 50.1\\
				ProDA Oracle~\cite{proda}&CLIP (B/16)*&RGB+SSG&71.1\\
				\textbf{KDA Oracle (Ours)}&CLIP (B/16)*&RGB+SSG&\textbf{73.2}\\
				\bottomrule
			\end{tabular}	
		}
		\vspace{-5pt}
	\end{table}
	
	\subsection{Compared with State-of-the-Art Methods}
	We evaluate the proposed method against state-of-the-art (SoTA) approaches on the Charades and SportsHHI datasets. On Charades, we report results under both the standard setting (predicted SSGs) and the oracle setting (ground-truth SSGs). On SportsHHI, we progressively integrate our KIM and KDM into baselines, and also train parameter-matched baselines for fair comparison.

	\textbf{Charades. }We compare the action recognition performance of the proposed method with state-of-the-art approaches on Charades, as shown in Table~\ref{table:charades}. Under both the standard and oracle settings, disentanglement-based methods (i.e., ours and ProDA~\cite{proda}) achieve the best performance, demonstrating their effectiveness in modeling object interactions within structured representations such as SSGs. Under the oracle setting, our method outperforms ProDA, indicating that incorporating atomic actions provides more effective semantic guidance for feature disentanglement. Under the standard setting, our method achieves performance comparable to ProDA, showing robustness under noisy SSG conditions. We further observe that the performance gap between the two settings is partly influenced by the feature space discrepancy between textual and visual representations. Specifically, the textual features are encoded by CLIP (B/16)~\cite{clip}, while the visual backbone is based on ResNet-101, leading to potential cross-modal misalignment. This mismatch may limit the effectiveness of knowledge transfer when ground-truth SSGs are unavailable, highlighting the importance of SSG quality for knowledge-guided disentanglement. Furthermore, when using the same backbone (CLIP B/16)~\cite{clip}, our method outperforms VicTR~\cite{victr}. This improvement is attributed to two factors: (1) the use of SSGs to capture structured object interactions, and (2) the incorporation of atomic actions with rich spatial-temporal information. Notably, these gains are achieved without introducing additional external models.

	\begin{table}
		\centering	
		\caption{Human–Human Interaction classification (HHICls) results on SportsHHI. ${\star}$ denotes our reimplementation based on the official code~\cite{sportshhi} with VideoMAE\cite{tong2022videomae}, with minor modifications (e.g., parameter size) to match the capacity of our method for a fair comparison. S and V denote SlowFast-R50~\cite{slowfast} and VideoMAE backbone, respectively.}
		\label{table:HHICls}
		\resizebox{\linewidth}{!}{
			\begin{tabular}{lllll}
				\toprule
				Method& mAP &R@50& R@20&Params\\
				\midrule
				STTran~\cite{cong2021spatial}&3.31&42.67&22.14&-\\
				HORT~\cite{hort}&3.75&50.33&26.96&-\\
				
				ACARN~\cite{acarn}&5.44&56.53&31.77&-\\
				
				
				\midrule
				SlowFast~\cite{slowfast}&5.00&52.74&26.82&48.11M\\
				\hspace{1em}+ KIM&6.96~\textcolor{red}{\small +1.96}&62.00~\textcolor{red}{\small +9.26}&32.38~\textcolor{red}{\small +5.56}&63.67M\\
				\midrule
				SportsHHI (S)~\cite{sportshhi}&7.52&59.53&32.76&76.52M\\
				\hspace{1em}+ KIM&7.48~\textcolor{green}{\small -0.04}&60.69~\textcolor{red}{\small +1.16}&28.88~\textcolor{red}{\small +3.88}&82.95M\\
				\hspace{1em}+ KIM \& KDM&8.77~\textcolor{red}{\small +1.25}&64.04~\textcolor{red}{\small +4.51}&36.83~\textcolor{red}{\small +4.07}&109.78M\\
				\midrule
				SportsHHI (V)$^{\star}$&10.20&67.12&36.01&131.44M\\
				SportsHHI (V)~\cite{proda}&10.16&68.40&40.16&123.11M\\
				SportsHHI (V)~\cite{sportshhi}&10.69&68.13&43.72&111.18M\\
				\hspace{1em}+ KIM&10.95~\textcolor{red}{\small +0.26}&71.35~\textcolor{red}{\small +3.22}&44.26~\textcolor{red}{\small +0.54}&124.45M\\
				\hspace{1em}+ KIM \& KDM&11.70~\textcolor{red}{\small +1.01}&69.14~\textcolor{red}{\small +1.01}&48.52~\textcolor{red}{\small +4.80}&134.48M\\
				\bottomrule
			\end{tabular}
		}
		\vspace{-5pt}
	\end{table}
	
	\textbf{SportsHHI. }We progressively incorporate KIM and KDM into different baselines, including SlowFast~\cite{slowfast} and SportsHHI~\cite{sportshhi} (Table~\ref{table:HHICls}). For SlowFast~\cite{slowfast}, we only add KIM since the original architecture lacks a disentanglement mechanism. The KIM-augmented method consistently improves all metrics, demonstrating strong generality and plug-and-play capability across backbones. When applied to SportsHHI~\cite{sportshhi}, KIM further improves performance, especially in Recall (e.g., R@50 and R@20), indicating that atomic action knowledge enhances feature discriminability. The gain in mAP is moderate, as such guidance activates multiple action components, improving coverage but introducing ambiguity in ranking. With KDM, the model achieves the best overall performance, showing that KIM enhances representations while KDM further disentangles them via semantic guidance, enabling clearer disentanglement and more structured modeling of entangled actions. Under comparable parameter settings, our methods outperform all baselines across metrics, showing the effectiveness of the proposed components.
	\subsection{Ablation Study}
	In this section, we first perform ablation studies on different configurations of KIM, KDM, and KD Loss. We then analyze the impact of different knowledge types. Finally, we evaluate the generalization ability of our method.
	
	\textbf{Impact of KIM and KDM. }KIM and KDM are key components of our framework. The former injects atomic action knowledge into video features, while the latter disentangles video features under more fine-grained knowledge guidance. As shown in Table~\ref{table:loss_ablation}, when KIM is removed (\ding{182}~\emph{v.s.}~\ding{183}), using KDM without knowledge guidance performs better than the variant with knowledge guidance. This can be attributed to the fact that the original features lack fine-grained information without the enhancement provided by KIM, making strong knowledge guidance prone to introducing misleading signals. In contrast, when KIM is included (\ding{184}~\emph{v.s.}~\ding{185}), KDM without knowledge becomes a limitation. Without knowledge guidance, learnable prompts alone struggle to provide effective guidance, especially when the test set contains actions that do not appear in the training set. By incorporating atomic action knowledge, KDM can better guide the disentanglement process. 
	
	\begin{table}
		\centering
		\caption{Ablation study of KIM, KDM, and KD Loss on Charades. LE and K denote learnable embeddings and knowledge, respectively.}
		\label{table:loss_ablation}
		\centering
		\resizebox{\linewidth}{!}{
			\begin{tabular}{ccccccc}
				\toprule
				&\makecell{KIM} 
				&\makecell{KDM w/ \\ LE} 
				&\makecell{KDM w/ \\ LE \& K} 
				&\makecell{$\mathcal{L}_{kd}$ w/ \\ margin} 
				&\makecell{$\mathcal{L}_{kd}$ w/o\\ margin} 
				&\makecell{${a}_{m}$ mAP on\\Charades (\%)} \\
				
				\midrule
				\ding{182}&-&\checkmark&-&-&-&72.48\\
				\ding{183}&-&-&\checkmark&-&-&71.97\\
				\ding{184}&\checkmark&\checkmark&-&-&-&68.11\\
				\ding{185}&\checkmark&-&\checkmark&-&-&72.85\\
				\ding{186}&-&\checkmark&-&\checkmark&-&72.40\\
				
				\ding{187}&\checkmark&-&\checkmark&-&\checkmark&72.92\\
				\ding{188}&\checkmark&-&\checkmark&\checkmark&-&\textbf{73.17}\\
				\bottomrule
			\end{tabular}	
		}
		\vspace{-5pt}
	\end{table}

	\textbf{Impact of KD Loss. }KD Loss guides the disentanglement process in KDM by introducing semantic knowledge cues, results are shown in Table~\ref{table:loss_ablation}. To preserve fine-grained information, we further apply margin-based clipping to similarity scores. Without knowledge guidance, introducing KD Loss even degrades performance (\ding{182}~\emph{v.s.}~\ding{186}), as the disentanglement constraint may separate features along arbitrary directions. In contrast, with semantic anchors, the model consistently improves (\ding{185}~\emph{v.s.}~\ding{187}), as disentanglement is aligned with meaningful action components. Moreover, the margin further stabilizes optimization by suppressing overly dominant similarities (\ding{187}~\emph{v.s.}~\ding{188}), enabling the model to capture finer-grained cues. Combining KIM, KDM, and KD Loss yields the best performance.
	
	\textbf{Impact of Different Knowledge Type. }We compare three types of knowledge: action labels, atomic actions only, and atomic actions combined with learnable embeddings. As shown in Table~\ref{table:loss_ablation_2}, removing the learnable embeddings in KDM leads to a significant performance drop (B~\emph{v.s.}~C). Using only atomic actions introduces overly strong semantic signals that may overwhelm video features and weaken the representation learning. Using labels as knowledge also results in inferior performance (A~\emph{v.s.}~C). Labels contain limited semantic information, and directly injecting label-level knowledge into SSG nodes may obscure finer-grained node information, thereby reducing the model’s representation capability. In contrast, combining atomic actions with learnable embeddings achieves the best performance by balancing semantic richness and flexibility.
	\begin{table}
		\caption{Ablation study of different knowledge types on Charades. ``LA", ``AT", and ``LE" denote action labels, atomic actions, and learnable embeddings, respectively.}
		\centering
		\label{table:loss_ablation_2}
		\resizebox{\linewidth}{!}{
			\begin{tabular}{ccccccc}
				\toprule
				&\makecell{KIM w/\\AT}&\makecell{KIM w/\\LA}&\makecell{KDM w/ \\ LE \& LA}
				&\makecell{KDM w/\\ AT} & \makecell{KDM w/ \\ LE\&AT}&\makecell{mAP on \\ Charades (\%)}\\
				\midrule
				A&-&\checkmark&\checkmark&-&-&70.96\\
				B&\checkmark&-&-&\checkmark&-&69.35\\
				C&\checkmark&-&-&-&\checkmark&\textbf{73.17}\\
				\bottomrule
				
			\end{tabular}
		}
	\end{table}
	
	\begin{table}
		\vspace{-5pt}
		\begin{minipage}{.48\linewidth}
			\caption{Ablation study of the domain shift on the Charades dataset. Acc, Std denote accuracy and standard deviation, respectively.}
			\resizebox{\textwidth}{!}{
				\label{table:shift}
				\begin{tabular}{ccc}
					\toprule
					Method&Acc&Std\\
					\midrule
					LaIAR~\cite{lair}&57.2&0.55\\
					ProDA~\cite{proda}&63.0&-\\
					Ours&65.1&1.67\\
					\bottomrule
				\end{tabular}
			}
		\end{minipage}
		\hfill
		\begin{minipage}{.48\linewidth}
			\caption{Comparison of accuracy using predicted and annotated relationships. N, G denote w/o, w/ ground-truth scene graph.}
			\resizebox{\textwidth}{!}{
				\label{table:relationship}
				\begin{tabular}{ccc}
					\toprule
					Method&N&G \\
					\midrule
					LaIAR~\cite{lair}&62.4&63.6\\
					ProDA~\cite{proda}&-&71.1\\
					Ours&70.9&73.2\\
					\bottomrule
				\end{tabular}
			}
		\end{minipage}
		\centering
		\vspace{-7pt}
	\end{table}

\captionsetup[table]{font=small}
\begin{table}
	\centering
	\caption{\small Head activation frequency (\%) under different KDM head numbers ($M$). Top indicates top-$k$\% routing weights. HI denotes head index. Empty cells are marked with `-'.}
	\label{table:head_activation}
	\vspace{-10pt}
	\resizebox{\linewidth}{!}{
		\begin{tabular}{c|c|cccccccccccc}
			\toprule
			$M$ & Top & HI0 & HI1 & HI2 & HI3 & HI4 & HI5 & HI6 & HI7 & HI8 & HI9 & HI10 & HI11 \\
			\midrule
			\multirow{2}{*}{8} 
			& 10\% & 0.00 & 0.00 & 0.00 & 62.49 & 0.00 & 6.26 & 31.25 & 0.00 & - & - & - & - \\
			& 50\% & 12.84 & 6.44 & 6.12 & 13.67 & 12.50 & 21.75 & 13.29 & 13.40 & - & - & - & - \\
			\midrule
			\multirow{2}{*}{10} 
			& 10\% & 25.22 & 0.00 & 0.00 & 0.00 & 0.00 & 0.00 & 25.00 & 0.00 & 24.78 & 25.00 & - & - \\
			& 50\% & 13.01 & 10.18 & 5.15 & 10.43 & 6.02 & 9.76 & 13.12 & 5.15 & 13.34 & 13.81 & - & - \\
			\midrule
			\multirow{2}{*}{12} 
			& 10\% & 0.00 & 0.00 & 16.71 & 25.38 & 0.00 & 0.00 & 22.99 & 0.00 & 0.00 & 0.00 & 34.93 & 0.00 \\
			& 50\% & 4.17 & 8.34 & 9.78 & 9.07 & 9.15 & 8.30 & 10.71 & 8.33 & 8.40 & 4.30 & 11.12 & 8.35 \\
			\bottomrule
		\end{tabular}
	}
	\vspace{-10pt}
\end{table}
	
	\textbf{Impact of Different Relationship. }We further analyze the impact of relationship information in scene graphs on our model. Specifically, we evaluate a variant that removes relationship information from the ground-truth scene graphs and relies only on object bounding boxes. LaIAR~\cite{lair} uses predicted scene graphs, while we do not use relationships. As shown in Table~\ref{table:relationship}, removing scene graphs results in a performance drop of 2.3 mAP, indicating that the spatial relationships provided by SSGs remain important for action understanding. Nevertheless, even without relationship information, our method achieves performance comparable to ProDA, suggesting that atomic actions can still provide useful relational cues even in the absence of explicit relationship annotations.
	
	\textbf{Impact of Different Domain. }RGB-based methods are sensitive to domain shifts in scene, viewpoint, and actors~\cite{zhang2022audio,lair,proda}. Following LaIAR, we split Charades into five non-overlapping scene subsets to evaluate robustness. As shown in Table~\ref{table:shift}, our method achieves the best average accuracy, demonstrating strong generalization. While exhibiting higher variance, this can be attributed to its overall performance and the sensitivity of semantic priors to scene changes. Importantly, it also attains the highest minimum accuracy, indicating robust worst-case behavior. Compared with ProDA~\cite{proda}, our method consistently outperforms under the same disentanglement setting, highlighting the benefit of atomic actions.
	
	\textbf{Impact of Disentanglement Heads Number.} We analyze how KDM allocates cues across heads by studying the routing weights under the 8/10/12-head settings (mAP: 71.14/73.17/73.25), as shown in Table~\ref{table:head_activation}. We observe that dominant semantic cues are mainly captured by a few core heads, while the remaining heads capture secondary semantic cues. In the 8-head setting, Top-10\% activations are highly concentrated (e.g., one head accounts for 62.49\%), indicating a capacity bottleneck. With 10/12 heads, high-weight cues are better distributed across core heads, and Top-50\% statistics show broader participation of non-core heads, explaining the improved performance. KD Loss further prevents semantic collapse and improves component diversity.

	\begin{figure}[!]
		\centering
		\includegraphics[width=\linewidth]{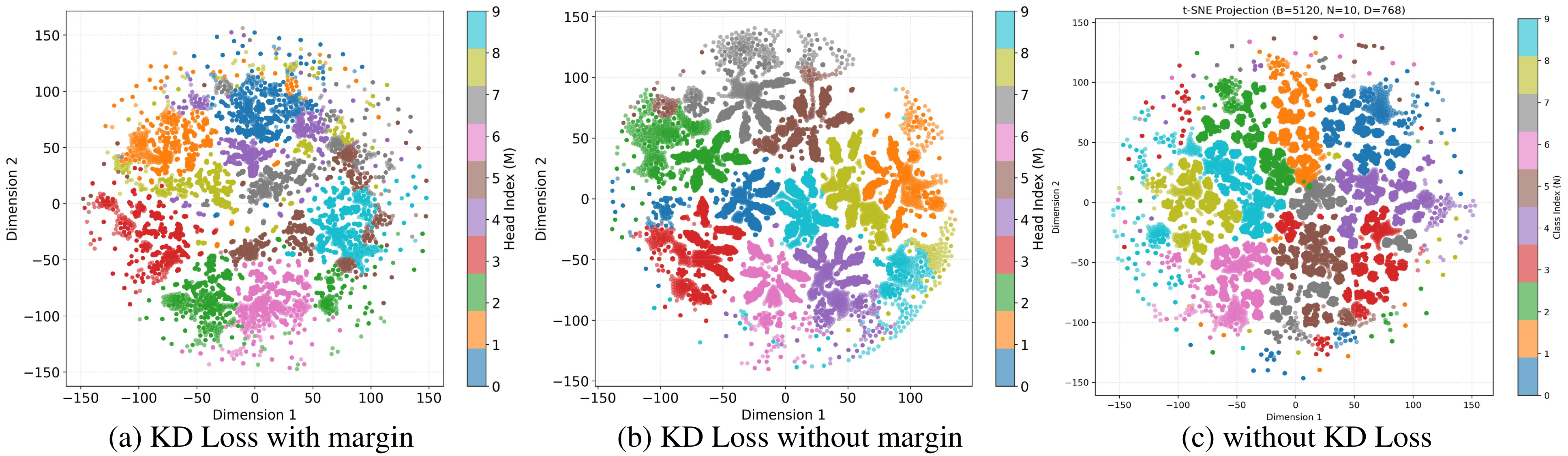}
		\caption{Visualization of disentangled knowledge in KDM under KD Loss with and without a margin constraint, using 10 disentanglement heads. (a) With margin. (b) Without margin. (c) Without KD Loss (i.e., no disentanglement regularization).}
		\label{fig:visual}
		\vspace{-15pt}
	\end{figure}
	
	\section{Qualitative Analysis}
	
In this section, we visualize and compare the learned feature distributions under three settings to investigate the effects of KD Loss and the margin constraint. Specifically, Fig.~\ref{fig:visual} presents the results with KD Loss and margin, with KD Loss but without margin, and without KD Loss.

With the margin constraint (Fig.~\ref{fig:visual}(a)), the features form compact yet naturally dispersed clusters. The margin acts as an implicit noise-filtering mechanism by allowing semantically weak components to remain unaligned rather than forcing them toward semantic anchors. This prevents noisy features from contaminating the anchors while preserving sufficient flexibility for clear semantic separation.

Without the margin (Fig.~\ref{fig:visual}(b)), all features are strictly aligned with the semantic anchors regardless of their semantic reliability. This leads to over-alignment, where noisy or weak components are also collapsed toward the anchors. Consequently, a single head may contain multiple sub-clusters, blurring semantic boundaries and reducing the expressiveness of the disentangled features.

Without KD Loss (Fig.~\ref{fig:visual}(c)), the features still exhibit coarse head-wise clustering, suggesting that part of the structure naturally emerges from the data distribution. However, the clusters are considerably more dispersed and contain more scattered points due to the absence of explicit supervision for semantic alignment and inter-head diversity. This results in poor intra-cluster compactness and weak inter-cluster separation.

Overall, KD Loss provides essential structural supervision by organizing the feature space into semantically meaningful clusters, while the margin constraint prevents excessive alignment and suppresses semantically weak components. Together, they produce more compact, well-separated, and expressive representations for knowledge disentanglement.

	\section{Conclusion}
	In this paper, we propose a novel framework for action disentanglement in multi-action scenarios by leveraging atomic actions with rich spatial-temporal semantics. We first decompose action labels into atomic action representations and organize them into memory banks, enabling effective knowledge modeling. Based on this, we introduce a Knowledge Injection Module (KIM) to enhance video representations through adaptive knowledge integration, and a Knowledge Disentanglement Module (KDM) to generate compact semantic guidance for disentangling specified actions. Conditioned on the Action Specification (AS), our method is able to disentangle any specified actions more precisely from complex scenes. Extensive experiments on benchmark datasets demonstrate that our approach achieves state-of-the-art performance and exhibits strong robustness. Moreover, our framework can be seamlessly integrated into different baselines, consistently improving performance, validating the effectiveness and generality of atomic knowledge for action understanding and disentanglement.

\section{Acknowledge}
This work was supported by grants from the Natural Science Foundation of Shaanxi Province (2024JCJCQN-66).

	\bibliographystyle{ACM-Reference-Format}
	\bibliography{sample-base}
	
	\clearpage
	\appendix
For better understanding of this work, we offer additional details, analysis, and results as follow:
\begin{itemize}
	\item \textbf{A. Details of Action Specification (AS). }In this section, we delve into the detailed construction of the Action Specification (AS) and present several representative examples to illustrate its formulation.
	\item \textbf{B. Analysis of Different Disentanglement Heads. } In this section, we provide detailed analyses of the selection strategy and ablation studies on the number of disentanglement heads.
	\item \textbf{C. More details of Qualitative Results. }	In this section, we provide additional qualitative analysis of KDM.
	\item \textbf{D. More details of Experiments.} In this section, we provide additional experimental details, including implementation specifics and the dataset splitting protocol used for domain shift evaluation.

	\item \textbf{E. Analysis of Atomic Action Generation.}
	In this section, we investigate the effects of different large language models and prompt templates on the generation of atomic action descriptions.
	
	\item \textbf{F. Efficiency Analysis.}
	In this section, we evaluate the computational efficiency of the proposed method in terms of inference latency and storage overhead.
\end{itemize}

\section{Action Specification}
\subsection{Design}
\label{as_design}

Let $C$ denote the total number of action classes. For illustration, we assume $C=5$. 
Given a video with ground-truth label set $\{0,1\}$, its multi-hot representation is:
\begin{equation}
	\label{video_label}
	[1,1,0,0,0].
\end{equation}

Based on the ground-truth labels, we first enumerate all non-empty subsets of the label set. For the example $\{0,1\}$, this results in three subsets: $\{0,1\}$, $\{0\}$, and $\{1\}$. Each subset is then converted into a \textbf{Specified Action Prompt (SAP)}, yielding three SAPs:
\begin{equation}
	\label{sap_example}
	\begin{aligned}
		&\{0,1\}\rightarrow[1,1,0,0,0], \\
		&\{0\}\rightarrow[1,0,0,0,0], \\
		&\{1\}\rightarrow[0,1,0,0,0].
	\end{aligned}
\end{equation}

However, using only SAPs may lead the model to rely excessively on label priors rather than visual content. 
To alleviate this issue, we introduce \textbf{distractor classes}, i.e., labels absent in the video. 
In this example, the distractor set is $\{2,3,4\}$. 

We fix the length of each SAP to $K$, which includes both present labels and distractor labels. 
Assuming $K=3$, we take the specified action corresponding to $\{0\}$ as an example. 
By injecting distractors, we obtain the following \textbf{distractor-injected SAPs}:
\begin{equation}
	\label{disap}
	\begin{aligned}
		&[1,0,1,1,0], \\
		&[1,0,1,0,1], \\
		&[1,0,0,1,1].
	\end{aligned}
\end{equation}

Similarly, for the specified actions $\{0,1\}$, we obtain:
\begin{equation}
	\label{disap2}
	\begin{aligned}
		&[1,1,1,0,0], \\
		&[1,1,0,1,0], \\
		&[1,1,0,0,1].
	\end{aligned}
\end{equation}

In both cases, each SAP contains exactly three positive entries.

\noindent
\textbf{Unspecified Action Prompt (UAP).}
We divide the feature space into two parts: one corresponding to the specified actions and the other to the remaining actions. 
Directly feeding SAP into the latter would cause information leakage, since SAP contains ground-truth labels. 

To avoid this, we construct \textbf{Unspecified Action Prompts (UAPs)} as the complement of SAP in the full label space, such that SAP and UAP together cover all action categories.

For Eq.~\ref{disap}, the corresponding UAPs are:
\begin{equation}
	\label{uap}
	\begin{aligned}
		&[0,1,0,0,1], \\
		&[0,1,0,1,0], \\
		&[0,1,1,0,0].
	\end{aligned}
\end{equation}

For Eq.~\ref{disap2}, the UAPs are:
\begin{equation}
	\label{uap2}
	\begin{aligned}
		&[0,0,0,1,1], \\
		&[0,0,1,0,1], \\
		&[0,0,1,1,0].
	\end{aligned}
\end{equation}

\begin{figure}[!t]
	\centering
	\includegraphics[width=1.00\linewidth]{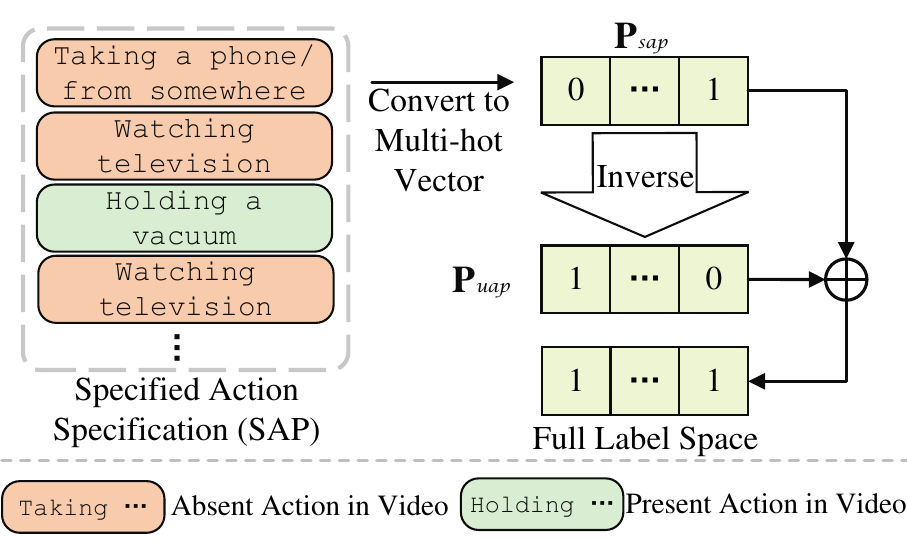}
	\caption{An illustration of action specification (AS) construction. The SAP, encompassing both present and absent actions in the video, is first mapped into the label space as a multi-hot vector $\mathbf{p}_{sap}$. Its logical complement yields $\mathbf{p}_{uap}$, ensuring that their sum entirely covers the complete label space.}
	\label{fig:as_construction}
	\label{fig:aamb}
	
\end{figure}

\subsection{Construction}
\label{as_csts}

The Action Specification (AS), defined as a SAP--UAP pair, consists of two types:
\textbf{(1) non-distractor-injected} and \textbf{(2) distractor-injected}.
In this section, we illustrate the construction process using a real video example from the Charades dataset, where the total number of action classes is $C=157$. For clarity, we represent SAPs and UAPs using index lists in this section due to the large label space ($C=157$). These index-based representations can be equivalently converted into multi-hot vectors following Eq.~\ref{sap_example}.

Given a video with ground-truth label set 
$\{77,79,80,76,75\}$, present labels are sampled from this set, 
while absent labels are sampled from the remaining label space 
$\{0,1,\dots,156\} \setminus \{77,79,80,76,75\}$.

\noindent \textbf{Non-distractor-injected. }Each SAP contains only present labels. We randomly sample subsets of size $1$ to $5$, generating:
\begin{equation}
	\label{non-dis}
	\begin{aligned}
		&\mathbf{p}_{non-sap}^{1}: \{77\}, \\
		&\mathbf{p}_{non-sap}^{2}: \{80, 76\}, \\
		&\mathbf{p}_{non-sap}^{3}: \{77, 80, 75\}, \\
		&\mathbf{p}_{non-sap}^{4}: \{79, 77, 76, 75\}, \\
		&\mathbf{p}_{non-sap}^{5}: \{80, 77, 76, 75, 79\}.
	\end{aligned}
\end{equation}

To prevent information leakage, the corresponding UAP is defined as the complement of each SAP in the full label space $\{0,\dots,156\}$:
\begin{equation}
	\label{non-dis_uap}
	\begin{aligned}
		&\mathbf{p}_{non-uap}^{1}: \underbrace{\{\textbf{80,76,75,79}, 0,1,\dots,156\}}_{\text{all labels except } \mathbf{p}_{non-sap}^{1}}, \\
		&\mathbf{p}_{non-uap}^{2}: \underbrace{\{\textbf{77,79,75}, 0,1,\dots,156\}}_{\text{all labels except } \mathbf{p}_{non-sap}^{2}}, \\
		&\mathbf{p}_{non-uap}^{3}: \underbrace{\{\textbf{79,76}, 0,1,\dots,156\}}_{\text{all labels except } \mathbf{p}_{non-sap}^{3}}, \\
		&\mathbf{p}_{non-uap}^{4}: \underbrace{\{\textbf{80}, 0,1,\dots,156\}}_{\text{all labels except } \mathbf{p}_{non-sap}^{4}}, \\
		&\mathbf{p}_{non-uap}^{5}: \underbrace{\{0,1,\dots,156\}}_{\text{all labels except } \mathbf{p}_{non-sap}^{5}}.
	\end{aligned}
\end{equation}

Here, the bold entries highlight the ground-truth labels that are excluded from the corresponding SAP, while each UAP contains all remaining labels in the full label space.

\noindent \textbf{Distractor-injected AS.}
We fix the SAP length to $K=16$. Each SAP contains both present and absent labels, where $0$ to $5$ labels are sampled from the ground-truth set, and the remaining labels are sampled from absent classes. This results in six SAPs:
\begin{equation}\small
	\label{dis}
	\begin{aligned}
		&\mathbf{p}_{dis-sap}^{1}: \{152, 32, 24, 125, 145, 110, 93, 138, 129, 44, 124, 130, 85, 7, 5, 0\}, \\
		&\mathbf{p}_{dis-sap}^{2}: \{\textbf{76}, 143, 111, 19, 104, 23, 112, 12, 78, 146, 56, 136, 116, 9, 24, 125\}, \\
		&\mathbf{p}_{dis-sap}^{3}: \{\textbf{77, 76}, 140, 111, 134, 130, 23, 18, 155, 136, 88, 133, 112, 127, 17, 2\}, \\
		&\mathbf{p}_{dis-sap}^{4}: \{\textbf{76, 79, 75}, 47, 125, 91, 71, 39, 66, 141, 9, 27, 52, 1, 132, 54\}, \\
		&\mathbf{p}_{dis-sap}^{5}: \{\textbf{76, 79, 75, 77}, 56, 88, 120, 22, 128, 62, 52, 105, 142, 124, 153, 35\}, \\
		&\mathbf{p}_{dis-sap}^{6}: \{\textbf{80, 77, 76, 75, 79}, 17, 51, 109, 143, 103, 57, 15, 152, 130, 125, 47\}.
	\end{aligned}
\end{equation}
the bold entries highlight the ground-truth labels.
To prevent information leakage, the corresponding UAP is defined as the complement of each SAP in the full label space $\{0,\dots,156\}$:
\begin{equation}\small
	\label{dis_uap}
	\begin{aligned}
		\mathbf{p}_{dis-uap}^{1}: & \underbrace{[\mathbf{80, 77, 76, 75, 79}, \, 0, 1, \dots, 156]}_{\text{all labels except } S_{dis-sap}^{1}}, \\
		\mathbf{p}_{dis-uap}^{2}: & \underbrace{[\mathbf{76, 79, 75, 77}, \, 0, 1, \dots, 156]}_{\text{all labels except }  S_{dis-sap}^{2}}, \\
		\mathbf{p}_{dis-uap}^{3}: & \underbrace{[\mathbf{76, 79, 75}, \, 0, 1, \dots, 156]}_{\text{all labels except }  S_{dis-sap}^{3}}, \\
		\mathbf{p}_{dis-uap}^{4}: & \underbrace{[\mathbf{77, 76}, \, 0, 1, \dots, 156]}_{\text{all labels except }  S_{dis-sap}^{4}}, \\
		\mathbf{p}_{dis-uap}^{5}: & \underbrace{[\mathbf{76}, \, 0, 1, \dots, 156]}_{\text{all labels } except S_{dis-sap}^{5}}, \\
		\mathbf{p}_{dis-uap}^{6}: & \underbrace{[0, 1, \dots, 156]}_{\text{all labels except }  \mathbf{p}_{dis-sap}^{6}}.
	\end{aligned}
\end{equation}

Here, bold entries indicate the ground-truth labels excluded from the corresponding SAP, while each UAP contains all remaining labels in the full label space. 
Finally, all SAPs and UAPs are converted into multi-hot vectors, resulting in
$\mathbf{p}_{sap} = \{\mathbf{p}_{non-sap}^{i}\}_{i=1}^{5} \cup \{\mathbf{p}_{dis-sap}^{i}\}_{i=1}^{6}$ 
and 
$\mathbf{p}_{uap} = \{\mathbf{p}_{non-uap}^{i}\}_{i=1}^{5} \cup \{\mathbf{p}_{dis-uap}^{i}\}_{i=1}^{6}$,
which are used as inputs to the model.

For a video with $N$ ground-truth labels, we generate a total of $2N+1$ SAP--UAP pairs. 
Specifically, the non-distractor-injected setting produces $N$ SAPs with $1$ to $N$ present labels (and their corresponding UAPs), while the distractor-injected setting produces $N+1$ SAPs with $0$ to $N$ present labels (and their corresponding UAPs).

\section{Analysis of Disentanglement Heads}

\noindent \textbf{Recap of KDM.}
The Knowledge Disentanglement Module (KDM) performs prompt-conditioned knowledge disentanglement guided by the Action Specification (AS).

Given a node feature $f_i^j$, KDM first retrieves action-relevant knowledge $\hat{z}$ conditioned on the AS, and decomposes it into $M$ \textbf{disentangled components} $\{h_m\}_{m=1}^M$ via $M$ parallel heads, each implemented as an MLP:
\begin{equation}\small
	h_m = \psi_m(\hat{z}),
\end{equation}
where $\psi_m(\cdot)$ denotes the $m$-th disentanglement head. Each component is expected to capture a distinct semantic factor.

These components are then dynamically aggregated using routing weights:
\begin{equation}\small
	\label{eq:guidance}
	g_{as} = \sum_{m=1}^{M} \alpha_m h_m, \quad
	\alpha = \mathtt{MLP}([f_i^j ; h_1 ; \dots ; h_M]),
\end{equation}
where $\alpha \in \mathbb{R}^{M}$ reflects the \textbf{head-wise importance} for each node, enabling adaptive selection of relevant components.

The resulting guidance signal is fused with the original node feature to produce the \textbf{guided representation}:
\begin{equation}\small
	\label{eq:fusion}
	\hat{f}_{i, as}^j = \mathtt{MLP}([f_i^j ; g_{as}]), \quad as \in \{SAP, UAP\}.
\end{equation}

By feeding $\mathbf{p}_{sap}$ and $\mathbf{p}_{uap}$ respectively, KDM produces two complementary guidance signals ($g_{sap}$ and $g_{uap}$), leading to two explicitly guided feature spaces ($\hat{f}_{i, sap}^j$ and $\hat{f}_{i, uap}^j$). These features are subsequently fed into the VGPNN for relational reasoning.

This design enables each node to selectively leverage disentangled knowledge components under different semantic prompts. Detailed analyses of the routing behavior and the number of disentanglement heads are provided in the supplementary material.

\noindent \textbf{Recap of KD Loss.}
The KD Loss regularizes the disentanglement heads by enforcing alignment to shared semantic anchors, while promoting diversity across heads. The semantic anchors are learnable prototypes that capture common action semantics.

Given the outputs of the $M$ heads $H = \{\hat{h}_1, \hat{h}_2, \dots, \hat{h}_M\}$ and a set of $L$ semantic anchors $\mathcal{U} = \{\mu_1, \mu_2, \dots, \mu_L\}$, the KD Loss is defined as:
\begin{equation}\small
	\mathcal{L}_{kd} = - \frac{1}{M} \sum_{m=1}^{M}
	\log
	\frac{ \sum_{l=1}^L \exp(s(\hat{h}_m, \mu_l) / t) }
	{ \sum_{l=1}^L \exp(s(\hat{h}*m, \mu_l) / t) + \sum_{n \neq m} \exp({\hat{h}_m}^\top \hat{h}_n / t) },
\end{equation}
where $t$ is the temperature parameter, and $s(\cdot)$ denotes a margin-bounded similarity function:
\begin{equation}\small
	s(\hat{h}_m, \mu_l) = \min(\hat{h}_m^\top \mu_l, \gamma).
\end{equation}

The numerator encourages each head to align with the shared semantic anchors, while the denominator additionally introduces competition among different heads, promoting inter-head diversity. The margin $\gamma$ bounds the similarity between heads and anchors, preventing trivial alignment and encouraging each head to retain distinctive information.

\noindent \textbf{Impact of Disentanglement Head Number.}
We study the effect of the number of disentanglement heads in KDM. 
As shown in Table~\ref{table:heads}, we jointly vary the number of heads $M$ and semantic anchors $L$ based on empirical settings. The performance first decreases as $M$ increases from $2$ to $8$, then improves and reaches the best result at $M=12$, and slightly drops at $M=14$. Although $M=12$ achieves the highest performance, the improvement over $M=10$ is marginal (73.25 \emph{vs.} 73.17). Meanwhile, the computational cost increases noticeably with $M$, since the KD loss involves pairwise similarity computation among disentangled components, resulting in $O(M^2)$ complexity (e.g., $10^2=100$ vs. $12^2=144$). Considering this trade-off between performance and efficiency, we adopt $M=10$ in our final model.

To better understand this behavior, we analyze the dynamic routing weights $\alpha$ under the settings of $M=8, 10, 12$. Specifically, we compute the activation frequency of each head when its weight $\alpha_m$ falls within the top 10\% to top 50\% across all samples.

As shown in Table~\ref{table:topk_8heads}, \ref{table:topk_10heads}, and \ref{table:topk_12heads}, we observe distinct routing patterns corresponding to different capacity regimes:

\textbf{1) Under-capacity ($M \leq 8$):} 
When $M=8$, the routing distribution is highly imbalanced. For example, a single head dominates the Top 10\% activations (62.49\%), while most other heads are rarely selected. This indicates a capacity bottleneck, where multiple semantic factors are forced to share limited representational components, leading to strong competition and suboptimal disentanglement.

\textbf{2) Balanced capacity ($M \in [10, 12]$):} 
When $M=10$ or $12$, the routing distribution becomes more structured. Under a strict threshold (Top 10\%), only a subset of heads is frequently activated, suggesting selective specialization. As the threshold is relaxed (Top 50\%), all heads participate more evenly, indicating sufficient coverage of diverse patterns. This balance between selective activation and broad participation suggests that the model can effectively allocate representational capacity to both dominant and fine-grained semantics, resulting in improved performance.

\textbf{3) Over-capacity ($M > 12$):} 
When further increasing $M$ (e.g., $M=14$), the performance slightly degrades. This may be attributed to over-parameterization, where the routing distribution becomes more diffuse and some heads are under-utilized. Such redundancy can weaken the discriminative power of individual components and reduce overall efficiency.

Overall, these results suggest that an appropriate number of disentanglement heads is critical for achieving a good balance between representation capacity and effective utilization.

\begin{table}
	\caption{Ablation study of different numbers of disentanglement heads in KDM. We accordingly adjust the number of learnable semantic anchors to match the number of heads.}
	\centering
	\label{table:heads}
	\resizebox{\linewidth}{!}{
		\begin{tabular}{cccccccc}
			\toprule
			
			Anchors Number (\textit{L})&16&32&48&56&64&72&80\\
			\midrule
			Heads Number (\textit{M})&2&4&6&8&10&12&14\\
			\midrule
			mAP on Charades&72.72&72.47&72.35&71.14&73.17&73.25&72.37\\
			\bottomrule
		\end{tabular}
	}
	
\end{table}

\begin{table}
	\caption{Head activation frequencies (\%) when their corresponding weights ($\alpha_m$) rank in the top 10\% to top 50\% across all samples ($M=8$). HI denotes the head index.}
	\centering
	\label{table:topk_8heads}
	\resizebox{\linewidth}{!}{
		\begin{tabular}{ccccccccc}
			\toprule
			HI & 0 & 1 & 2 & 3 & 4 & 5 & 6 & 7 \\
			\midrule
			10\% & 0.00 & 0.00 & 0.00 & 62.49 & 0.00 & 6.26 & 31.25 & 0.00 \\
			20\% & 6.26 & 0.00 & 0.00 & 31.25 & 0.00 & 31.25 & 15.62 & 15.62 \\
			30\% & 20.83 & 7.68 & 0.00 & 20.83 & 0.00 & 20.83 & 19.39 & 10.43 \\
			40\% & 15.73 & 7.83 & 0.31 & 16.00 & 9.69 & 19.05 & 15.73 & 15.66 \\
			50\% & 12.84 & 6.44 & 6.12 & 13.67 & 12.50 & 21.75 & 13.29 & 13.40 \\
			\bottomrule
		\end{tabular}
	}
\end{table}

\begin{table}
	\caption{Head activation frequencies (\%) when their corresponding weights ($\alpha_m$) rank in the top 10\% to top 50\% across all samples ($M=10$). HI denotes the head index.}
	\centering
	\label{table:topk_10heads}
	\resizebox{\linewidth}{!}{
		\begin{tabular}{ccccccccccc}
			\toprule
			HI&0&1&2&3&4&5&6&7&8&9\\
			\midrule
			10\%&25.22&0&0&0&0&0&25.00&0&24.78&25.00\\
			20\%&25.00&0&0&12.36&6.71&0&24.91&5.89&12.63&12.50\\
			30\%&16.67&8.33&0&8.33&8.33&0.27&16.67&8.33&16.67&16.40\\
			40\%&13.37&12.509&0.01&12.50&6.29&7.73&13.99&6.25&13.48&13.87\\
			50\%&13.01&10.18&5.15&10.43&6.02&9.76&13.12&5.15&13.34&13.81\\
			\bottomrule
		\end{tabular}
	}
	
\end{table}
\begin{table}
	\caption{Head activation frequencies (\%) when their corresponding weights ($\alpha_m$) rank in the top 10\% to top 50\% across all samples ($M=12$). HI denotes the head index.}
	\centering
	\label{table:topk_12heads}
	\resizebox{\linewidth}{!}{
		\begin{tabular}{ccccccccccccc}
			\toprule
			HI & 0 & 1 & 2 & 3 & 4 & 5 & 6 & 7 & 8 & 9 & 10 & 11 \\
			\midrule
			10\% & 0.00 & 0.00 & 16.71 & 25.38 & 0.00 & 0.00 & 22.99 & 0.00 & 0.00 & 0.00 & 34.93 & 0.00 \\
			20\% & 0.00 & 0.00 & 16.46 & 20.83 & 20.83 & 0.00 & 20.83 & 0.00 & 0.21 & 0.00 & 20.83 & 0.00 \\
			30\% & 0.00 & 13.18 & 13.89 & 13.89 & 13.89 & 0.00 & 13.90 & 0.17 & 10.22 & 6.94 & 13.92 & 0.00 \\
			40\% & 5.21 & 10.42 & 10.44 & 10.43 & 10.42 & 5.21 & 10.62 & 5.51 & 10.42 & 5.21 & 10.91 & 5.21 \\
			50\% & 4.17 & 8.34 & 9.78 & 9.07 & 9.15 & 8.30 & 10.71 & 8.33 & 8.40 & 4.30 & 11.12 & 8.35 \\
			\bottomrule
		\end{tabular}
	}
\end{table}

\begin{figure*}[!t]
	\centering
	\includegraphics[width=\linewidth]{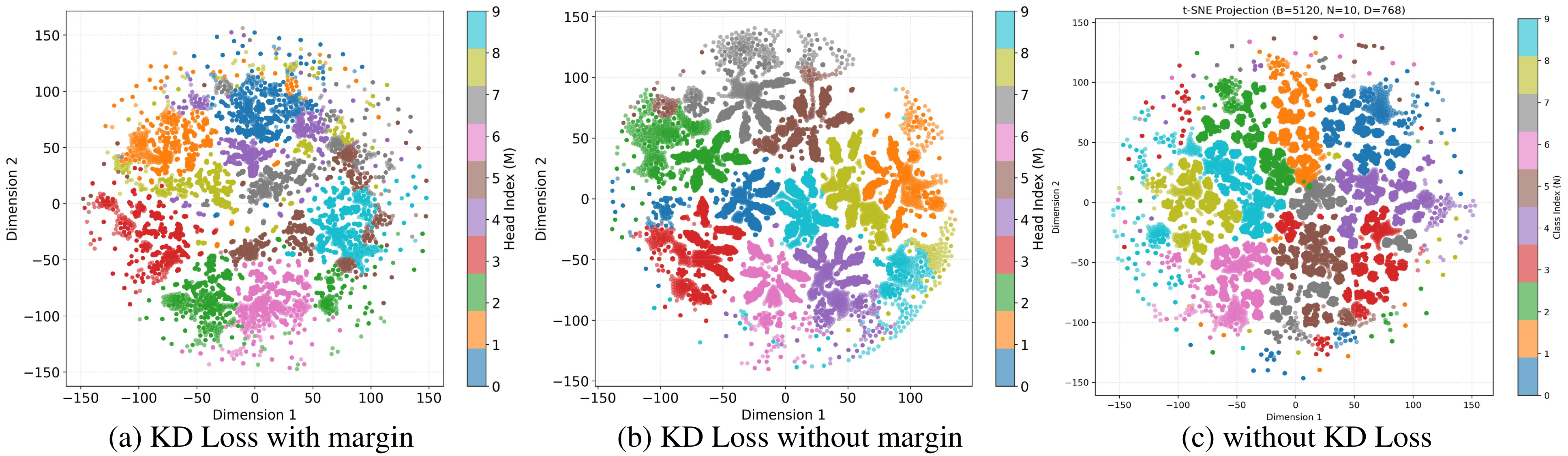}
	\caption{Visualization of disentangled knowledge in KDM under KD Loss with and without a margin constraint, using 10 disentanglement heads. (a) With margin. (b) Without margin. (c) Without KD Loss (i.e., no disentanglement regularization).}
	\label{fig:visual2}
\end{figure*}

\section{Qualitative Results}

As shown in Fig.~\ref{fig:visual2}, feature distributions are fundamentally different under three settings: with KD Loss and margin, with KD Loss but without margin, and without KD Loss. 

With the margin constraint (Fig.~\ref{fig:visual2}(a)), representations exhibit more natural dispersion. This reflects a desirable relaxation that acts as an implicit \emph{noise-filtering mechanism}: semantically weak components are allowed to remain unaligned rather than being forcibly projected onto anchors. By preventing invalid features from contaminating the anchors, the margin preserves representational flexibility and enables clearer semantic separation.

In contrast, in the no-margin case (Fig.~\ref{fig:visual2}(b)), the lack of constraint forces all features, regardless of semantic validity, into \emph{strict alignment} with semantic anchors. This rigid normalization leads to severe \emph{over-alignment}: even noisy or semantically weak components are collapsed into anchors, where a single head may contain multiple sub-clusters within a shared space. This blurs semantic boundaries and reduces the expressiveness of disentangled features, ultimately degrading recognition performance.

Without KD Loss (Fig.~\ref{fig:visual2}(c)), the model lacks explicit supervision for both shared semantic alignment and inter-head diversity. Interestingly, the features still exhibit a coarse clustering structure, where each head forms a loosely grouped cluster composed of several sub-clusters. This pattern is similar to the no-margin case, suggesting that such clustering behavior naturally emerges from the data distribution even without anchor-based guidance. However, these clusters are significantly more dispersed and less compact compared to those with KD Loss. The absence of semantic anchors leads to weak alignment, causing features to spread excessively and resulting in poor intra-cluster compactness and inter-cluster separability. 

Moreover, a noticeable number of scattered points can be observed, indicating the presence of semantically weak or noisy components. This observation aligns with our analysis under KD Loss: such weak semantics inherently exist, but without proper regularization, they cannot be effectively controlled or separated. 

Overall, the visualization shows that KD Loss provides essential structural supervision by organizing the feature space into semantically meaningful clusters, while the margin constraint further refines this structure by balancing alignment and flexibility. Together, they enable more compact, well-separated, and expressive representations for disentangled knowledge.

\section{More details of Experiments}
For each video, we uniformly sample 16 frames as input. 
All experiments are conducted using two RTX 4090D GPUs. 
Unless otherwise specified, we follow standard training settings; for Charades, we use a learning rate of $1.5\times10^{-5}$, a batch size of 16, and train for 5 epochs.

\begin{table}[H]
	\centering
	\caption{ This table outlines the division of the Charades dataset into five subsets, ensuring that there is no overlap between the scenes used
		for training and testing}
	\label{subdataset}
	\resizebox{1.0\linewidth}{!}{
		\begin{tabular}{ccc}
			\toprule
			Subdataset&Training Scene& Test Scene\\
			\midrule
			Scenario1&\makecell{Stairs,Laundry room,Home Office,\\
				Hallway,Bedroom,Pantry,Dining room,Entryway}
			&\makecell{Living room,Closet,Kitchen,Bathroom,\\
				Garage,Recreation room,Basement,Other}\\
			\midrule
			Scenario2&\makecell{Laundry room,Bathroom,Pantry,Closet,\\
				Entryway,Recreation room,Garage,Other}&\makecell{Bedroom,Living room,Kitchen,Home Office,\\
				Hallway,Stairs,Basement,Dining room}\\
			\midrule
			Scenario3&\makecell{Stairs,Laundry room,Bedroom,Basement,\\
				Bathroom,Entryway,Recreation room,Other}&\makecell{Living room,Closet,Kitchen,Home Office,\\
				Garage,Hallway,Pantry,Dining room}\\
			\midrule
			Scenario4&\makecell{Kitchen,Stairs,Laundry room,Home Office,\\
				Bedroom,Bathroom,Pantry,Dining room}&\makecell{Living room,Closet,Garage,Hallway,\\
				Recreation room,Entryway,Basement,Other}\\
			\midrule
			Scenario5&\makecell{Kitchen,Laundry room,Hallway,Basement,\\
				Dining room,Living room,Closet,Other}&\makecell{Bedroom,Home Office,Bathroom,Garage,\\
				Stairs,Recreation room,Entryway,Pantry}\\
			\bottomrule
		\end{tabular}
	}
\end{table}

To evaluate domain shifts, we partition the Charades dataset into five subsets with disjoint training and testing scenes. The detailed splitting protocol is described here, while the corresponding performance is reported in the main paper (Table.~\ref{subdataset}).

\begin{table}

	\centering
	\caption{\small KDA performance with different LLMs and prompt types. ST denotes Spatial-temporal prompts which is used in the paper.}
	\label{table:LLM_Prompts}

	\centering
	\resizebox{\linewidth}{!}{
		\begin{tabular}{lllll}
			\toprule
			LLMs&ST&Motion&Object&Non-template\\
			\midrule
			Qwen3-plus&73.17&71.99&72.63&71.73\\
			GPT-5.5&73.30&73.16&72.74&72.31\\
			Opus-4.8&72.95&72.41&72.58&71.91\\
			\bottomrule
		\end{tabular}	
	}

\end{table}

\begin{table}
	\centering
	\caption{\small Efficiency under matched parameters.}
	\label{table:inference}
	\centering
	\resizebox{\linewidth}{!}{
		\begin{tabular}{lllll}
			\toprule
			Methods&Params (M)&GPU Memory (GiB)&Inference Latency (s)&mAP\\
			\midrule
			SportsHHI (re-implemented)&123.11&2.46&0.0305&10.16\\
			SportsHHI (re-implemented)&131.44&2.55&0.0307&10.20\\
			\midrule
			SportsHHI (baseline)&111.18&2.32&0.0298&10.69\\
			\hspace{1em}+ KIM&124.45&2.47&0.0320&10.95\\
			\hspace{1em}+ KIM \& KDM&134.48&2.58&0.0420&11.70\\
			\bottomrule
		\end{tabular}	
	}

\end{table}

\section{Analysis of Atomic Action Generation}
We evaluate atomic actions generated by different LLMs and prompt strategies in Table~\ref{table:LLM_Prompts}. Specifically, we consider three structured prompt templates and one unconstrained prompt. The \textbf{Spatial-temporal (ST) prompt} describes each action from a spatio-temporal perspective by first specifying the typical scene context and then decomposing the action into three temporally ordered atomic steps, thereby capturing both where the action occurs and how it evolves over time. The \textbf{Motion prompt} focuses primarily on observable motion patterns, such as human body movements, object movements, interaction dynamics, and state transitions, while reducing the emphasis on contextual scene information. In contrast, the \textbf{Object prompt} emphasizes the objects involved in the action, their semantic roles, human--object relations, and possible changes in object states during the interaction. The \textbf{non-template prompt} directly asks the LLM to decompose an action label without imposing an explicit output structure or predefined semantic perspective.

Comparable results across Qwen3-Plus, GPT-5.5, and Opus-4.8 indicate that KDA is not tied to a specific LLM backbone. Moreover, the three structured prompts consistently outperform the non-template prompt, demonstrating that explicit generation constraints help produce more stable and informative atomic-action descriptions. Although the structured templates emphasize different aspects of an action, their relatively similar performance suggests that KDA is robust to variations in prompt design and potential noise in LLM-generated knowledge. This robustness mainly stems from the adaptive retrieval mechanism in KDA, which selectively retrieves action-relevant atomic knowledge instead of directly injecting all generated descriptions into the visual representation. Among the three structured templates, the ST prompt adopted in the main paper achieves the best performance, indicating that jointly modeling scene context and temporal action evolution provides more comprehensive guidance for action recognition. All atomic-action descriptions are generated offline and fixed before model training; therefore, no LLM inference is required during either training or testing.

\section{Efficiency Analysis}
We report the inference latency and GPU memory consumption under parameter-matched settings in Table~\ref{table:inference}. Compared with the parameter-matched SportsHHI baseline (131.44M parameters), KDA improves the mAP from 10.20 to 11.70 while introducing only a marginal increase in GPU memory, from 2.55 to 2.58 GiB. In particular, KIM increases the inference latency only slightly, from 0.0298\,s to 0.0320\,s, demonstrating the efficiency of feature-level knowledge injection. The additional latency mainly arises from KDM, as action disentanglement requires two forward passes, whereas the baseline performs only one. Nevertheless, this computational overhead yields a clear improvement in recognition performance. Moreover, all atomic actions are generated once offline and fixed before training; therefore, no LLM inference is involved during either training or testing.

\end{document}